\documentclass[runningheads]{llncs}
\usepackage{graphicx}
\usepackage{amssymb}
\usepackage{booktabs}

\begin{document}
\title{An Evoked Potential-Guided Deep Learning Brain Representation For Visual Classification}
%
%
\author{Xianglin Zheng\inst{} \and
Zehong Cao\inst{} \and
Quan Bai\inst{}}
\authorrunning{X. Zheng et al.}
%
\institute{Discipline of ICT, University of Tasmania, Australia} 

\maketitle              
\begin{abstract}
The new perspective in visual classification aims to decode the feature representation of visual objects from human brain activities. 
Recording electroencephalogram (EEG) from the brain cortex has been seen as a prevalent approach to understand the cognition process of an image classification task. In this study, we proposed a deep learning framework guided by the visual evoked potentials, called the Event-Related Potential (ERP)-Long short-term memory (LSTM) framework, extracted by EEG signals for visual classification. In specific, we first extracted the ERP sequences from multiple EEG channels to response image stimuli-related information. Then, we trained an LSTM network to learn the feature representation space of visual objects for classification. In the experiment, 10 subjects were recorded by over 50,000 EEG trials from an image dataset with 6 categories, including a total of 72 exemplars. Our results showed that our proposed ERP-LSTM framework could achieve classification accuracies of cross-subject of 66.81\% and 27.08\% for categories (6 classes) and exemplars (72 classes), respectively. Our results outperformed that of using the existing visual classification frameworks, by improving classification accuracies in the range of 12.62\% - 53.99\%.  Our findings suggested that decoding visual evoked potentials from EEG signals is an effective strategy to learn discriminative brain representations for visual classification.

\end{abstract}

\section{Introduction}

Visual classification is a computer vision task that inputs an image and outputs a prediction of the category of the object image. It has become one of the core research directions of computer vision and been widely used in many applications such as localisation, detection, and segmentation of objects \cite{karpathy2016cs231n,He_2019_CVPR}. In the last decades, with the discovery of Convolutional Neural Networks (CNN) that is loosely inspired by the human visual architecture, researchers have made great efforts on performance improvement in object recognition \cite{Girshick_2014_CVPR,10.1007/978-3-319-10578-9_23,He_2017_ICCV}. However, some researchers are cognizant that there are still significant differences in the way that human and current CNN process visual object information \cite{NIPS2018_7982}. Particularly, the performance of which evaluated CNNs on negative images \cite{8260656} and compared generalisation towards previously unseen distortions in human and CNNs \cite{NIPS2018_7982} have further shown the robustness of CNNs on object recognition are not at the human level.

For human beings, the advantage of visual exteroceptive sense is distinct. For example, someone usually directly looks at the objects they want to recognise to make full use of the foveal vision. It has always been a challenging issue in cognitive neuroscience to figure out how humans accomplish the object recognition effortlessly in everyday life and how to model the mechanisms or the discriminative feature spaces employed for object categorisation \cite{Grill1988,Beeck10111,Peelen2007The}. Researchers have investigated that the brain exhibits functions of feature extraction, shape description, and memory matching, when the human brain is involving visual cognitive processes \cite{gazzaniga2006cognitive}. Subsequent studies \cite{Simanova_2010,NORMAN2006424,HANSON2004156} have further revealed that analysing brain activity recordings, linkage with the operating human visual system, is possible to help us understand the presentational patterns of visual objects in the cortex of the brain. For instance, from the hemodynamic responses, information of some particular categories of images (e.g., human faces and objects) were found to be presented primarily in several specialised regions in brain cortex \cite{Haynes2005,Ishai9379,COX2003261}. Inspired from the above visual neuroscience investigations, there is some recent work considered to process visual classification problems by analysing neurophysiology and neuroimaging signals recorded from human visual cognitive processes \cite{Haynes2006Decoding,Kaneshiro2015,spampinato2017,renli2018,renli2020}, which have demonstrated the feasibility to identify the feature space that employed by humans for object image categorisations. However, they are still limited to analyse the brain visual activities by using the raw physiological signals without extracting a more representative input during the signal preprocessing stage.

In addition, many existing visual classification studies have been focusing on electroencephalography (EEG)-based visual object discriminations as we explored above. EEG signals, featuring by a high temporal resolution in comparison with other neuroimaging, are generally recorded by electrodes on the surface of the scalp, which has been applied in developing several areas of brain-computer interface (BCI) classification systems, such as pictures, music, and speech recognitions \cite{bashivan2015learning,stober2015deep,Carlson_2011}. Interestingly, when human evoked by the different visual or auditory stimulus, the EEG signals, could collect diverse responses of evoked potentials \cite{gazzaniga2006cognitive}. However, the raw waveforms of EEG signals are the spontaneous potential of the human brain in a natural state, which is difficult to distinguish the hidden event-related information during the visual cognitive process \cite{gazzaniga2006cognitive,DEPASCALIS2004295}. Thus, the event-related potential (ERP) was proposed to identify the real-time evoked response waveforms caused by stimuli events (e.g., specific vision and motion activities), which usually performed lower values than the spontaneous EEG amplitude \cite{gazzaniga2006cognitive} and extracted from the EEG fragments with averaged superposition in multiple visual trials. 

\section{Related Work}

Decoding image object-related EEG signals for visual classification has been a long-sought objective. For example, the early-stage studies in \cite{philiastides2006neural,Philiastidesbhi130} attempted to classify single-trial EEG responses to photographs of faces and cars. An image classification task \cite{Kaneshiro2015} in 2015 considered a comprehensive linear classifier to tackle EEG brain signals evoked by 6 different object categories, and achieved the classification accuracy around 40\%. Then, investigating the intersection between deep learning and decoding human visual cognitive feature spaces has increased significantly. For example, Bashivan's work \cite{bashivan2015learning} in 2015 transformed EEG into topographical maps for each image and trained a deep recurrent-convolutional neural network to learn robust representations and classification. 

Afterwards, Sampinato et al. \cite{spampinato2017} proposed an automated visual classification framework in 2017 to compute EEG features with Recurrent Neural Networks (RNN) and trained a CNN-based regressor to project images onto the learned EEG features. However, the recent two studies in 2018 and 2020 \cite{renli2018,renli2020} brought force questions to Spampinato's block design \cite{spampinato2017} employed in the EEG data acquisition, where all stimulus of a specific class are presented together without randomly intermixed. In particular, the latest study in 2020 \cite{renli2020} replicated the Spampinato's experiment \cite{spampinato2017} with a rapid-event design and analysed the classification performance on the randomised EEG trials. In addition, we noted that a special structure recurrent neural network, Long Short-Term Memory (LSTM) network, is commonly used in these studies to learn the representations of brain signals, which have shown the feasibility to decode human visual activities and deep learning for visual classification. 

However, most of current machine learning approaches for visual classification ignored to explore the EEG evoked potentials of spontaneous generation. Even now deep learning is still difficult to recognise distinctive patterns of evoked potentials from the raw waveforms of EEG signals with a visual stimulus, so we assume that excluding visual related evoked potentials could be a fundamental cause that leads to an uncertain feature representation space for visual classification and place a restriction on the improvement of classification accuracy. Only two early-stage studies \cite{Wang_2012,qin2016classifying} were preliminarily explored evoked-guided models by extracting ERPs from EEG signals for visual classification and used 4 components of ERP (P1, N1, P2a, and P2b) as the inputs of a linear classifier, but these preliminary studies required multiple spatial or temporal components during the preprocessing stage and did not apply deep learning for evoke-guided feature representations. 

Thus, in this study, our work was inspired from two assumptions: (1) the feature representations employed by human brains for visual classification will be more pronounced learned from the purer ERP which conveys image stimuli-related information; (2) the multi-dimensional ERPs can be decoded to obtain a one-dimensional representation using RNN and do not require pre-selection of spatial or temporal components. One special type of RNNs, the LSTM, presents the strong capability in recognising long-term and short-term feature representations from time-series EEG signals.

With the above two assumptions, in this study, we proposed the first visual evoked potential-guided deep learning framework, called ERP-LSTM framework, to learn the discriminative representations for visual classification. The ERP-LSTM framework is constituted by two stages: (1) acquiring the ERP waveforms from multiple EEG trials with averaged superposition; (2) a parallel LSTM network mapping the extracted ERPs into feature representation vectors and involving an activation layer that classifies the derived vectors into different classes.

\section{Our Proposed Framework}

\begin{figure}
\includegraphics[width=\textwidth]{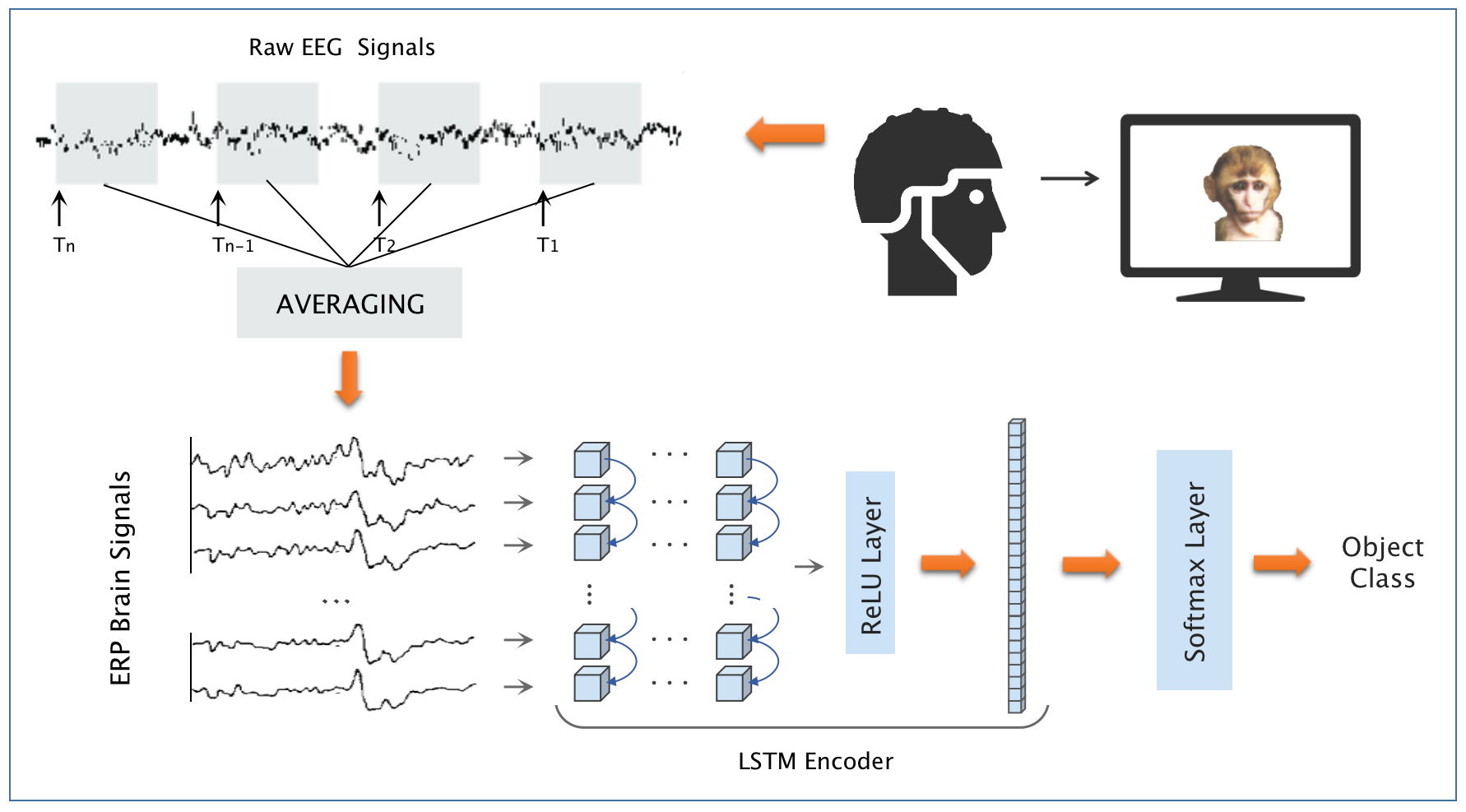}
\caption{The overview of the ERP-LSTM framework} \label{fig1}
\end{figure}

The overview of our proposed ERP-LSTM framework is shown in Fig.~\ref{fig1}, which is separated into two stages for visual classification. In Stage 1, we employed raw EEG signals recorded from the visual experiment and then extracted ERPs from the raw EEG data to secure the visual stimuli-related signals. In Stage 2, we trained an LSTM network to learn the representation space of the ERP sequences and followed a Softmax classification trained to discriminate the different classes of the images.

\subsection{Stage 1: ERPs Extractions from EEG}

\label{section: Stage 1: Extracted ERPs from raw EEG}
The representative features of EEG signals play an essential role in classifying image object categories. The first stage of our proposed framework aims to extract representative visual-related features of ERPs by increasing the signal-noise ratio (SNR) of the raw EEG signals with smooth-averaging measurement. A number of EEG segments with the same trials are averaged out to a fused waveform. In specific, during the averaging process, the consistent features of the segments (the ERPs) are retained, while features that vary across segments are attenuated (refer to the upper left corner of Fig.~\ref{fig1}).

More formally, let $d_{i}^{j}=\left\{T_{1}^{j}, T_{2}^{j}, \ldots, T_{n}^{j}\right\}$, $i \times n = N $, $d_{i}^{j}$ is the $i_{th}$ subset of the multi-channel temporal EEG signals, when one subject is viewing the $j_{th}$ exemplar image. N is the number of EEG trials to be averaged, which contains $n$ of EEG trials, where trial $T_{n}^{j} \in \mathbb{R}^{c}$ (c is the number of channels). 

The averaging process is described by the following fomula:

\begin{equation}
e_{i}^{j}=\left(\sum  T_{n}^{j}\right) / n, \quad T_{n}^{j} \in d_{i}^{j}
\end{equation}
where $e_{i}^{j}$ is the ERP sequence averaged from $d_{i}^{j}$. 

Let $E$ be the sum of extracted multi-channel ERPs, $E=\left\{e_{1}^{j}, e_{2}^{j}, \ldots, e_{i}^{j}\right\}$, which will be the inputs of the LSTM encoder module we addressed in the next subsection to learn discriminative feature representations for visual classification.

\subsection{Stage 2: Feature Representations and Classification}
\label{Stage 2: Learning representations and Classification}

To further utilise the spatial and temporal information from extracted ERPs, we applied an LSTM encoder module shown in the lower part of Fig.~\ref{fig1}, which refers to Spampinato's ``common LSTM + output layer'' architecture \cite{spampinato2017}. The inputs of the encoder are the multi-channel temporal signals - ERPs, which are preprocessed in the previous subsection. 

At each time step t, the first layer takes the input $s(·, t)$ (the vector of all channel values at time $t$), namely that all ERPs from multiple channels are initially fed into the same LSTM layer. After a stack of LSTM layers, a ReLU layer is added to make the encoded representations easy to map the feature space. The whole LSTM encoder outputs a one-dimensional representation feature of each ERP. After the representation vectors are obtained, a Softmax activation layer is finally connected to classify the LSTM representative features to different visual categories.

The LSTM encoder module is evaluated by the cross-entropy loss, which measures the differences between the classes predicted from the network and the ground-truth class labels. The cross-entropy loss $H(p, q)$ is defined as follows:
\begin{equation}
H(p, q)=-\sum_{x}(p(x) \log q(x)+(1-p(x)) \log (1-q(x)))
\end{equation}
where $p(x)$ is the probability distribution vector of the ground-truth class label, and $q(x)$ is the output probability distribution of the Softmax classification layer. The total loss is propagated back into the neural network to update the whole model's parameters through gradient desent optimisation.

In the proposed ERP-LSTM framework, the LSTM encoder module is used for generating feature representations from ERP sequences, followed by a Softmax classification layer to predict the visual classes.

\section{The Experiment}
\subsection{The Dataset}

\begin{figure}
\includegraphics[width=\textwidth]{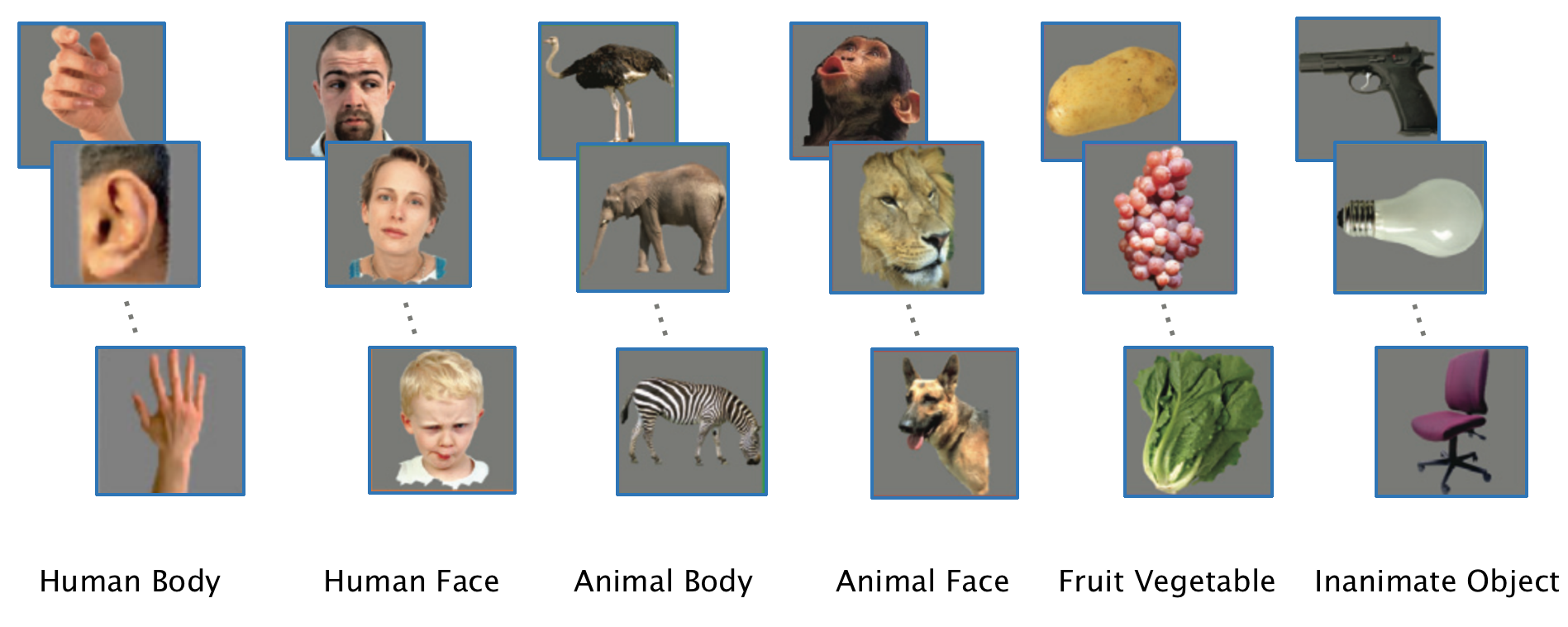}
\caption{Some samples of photographs with six categories} \label{fig2}
\end{figure}

In this study, we used an image dataset, along with EEG signals that were collected from 10 subjects viewed different categories of images \cite{Kaneshiro2015}. The image dataset has a total of 72 photographs containing 6 categories: Human Body (HB), Human Face (HF), Animal Body (AB), Animal Face (AF), Fruit Vegetable (FV), and Inanimate Object (IO),  As shown in Fig. \ref{fig2}, we demonstrated some sample of photographs from six categories (HB, HF, AB, AF, FV, and IO) that were used for visual classification.

During the experiment, each subject completed two experimental sessions, and each of the sessions contained three blocks. In each block, every image from all 72 images was displayed 12 times in random order. Each experiment trial consists of a single image, displayed 500 ms on the screen, followed by 750 ms a blank grey screen as an interval. In total, each participant completed 72 trials of each of the 72 images and conducted a total of 5,184 trials per subject. Thus, 51,840 trials in total were collected from 10 subjects. In this study, each of the trials was labelled to map the description of the visual evoked-related events, namely the corresponding image category (which of the 6 categories is related to) or image exemplar (which of the 72 images is related to).

In terms of EEG recordings, it was collected using unshielded 128-channel EGI HCGSN 110 nets, and preprocessed using a butter-worth filter with cut-off frequencies of 1 Hz and 25 Hz before temporally downsampled to a final sampling rate of 62.5 Hz (496 ms). The 4 channels at the forefront of the frontal cortex were deleted for artifact removal purpose so that we remained 124 channels electrodes for ERP analysis.

\subsection{Settings}
In this study, we randomly segmented the 72 EEG trials into 6 sets, and each set contains 12 EEG trials. The trials in each set are averaged to extract an ERP sequence with the same image and category label. Then, we obtained 6 ERP sequences of each image and also achieved $E$, the ERP space of the overall extracted 124-channel ERP sequences. Of note, the ERP space $E$ is split into the training set and the testing set with a proportion of 5:1, indicating that 80\% ERP sequences for each image keep in the training set and the remaining 20\% sequences are on the testing set. To further evaluate the performance of the classification framework, we performed two types of data classification: cross-subject and within-subject basis.

\section{Results}

\subsection{Performance of Six-category Visual Classification}

As shown in Table \ref{tabel 1}, we presented the classification performance of the basic LSTM using raw EEG (EEG-LSTM) \cite{spampinato2017} and our proposed ERP-LSTM frameworks. It also illustrated the two types (cross-subject and within-subject) of classification performance. Our findings showed that our proposed ERP-LSTM framework could reach about 66.81\% accuracy for cross-subject type of visual classification and achieve the highest classification accuracy of 89.06\% for a single subject (subject 1). Both outcomes were outperformed that of  EEG-LSTM framework, where the classification accuracy improved 30.09\% across 10 subjects, 53.99\% for subject 1, and 23.46\% for averaged within-subject from 1 to 10. 

\begin{center}
\begin{table}
\caption{Performance of six-category visual classification}
\setlength{\tabcolsep}{4mm}{
\begin{tabular}{cccc}
\toprule
Accuracy               & EEG-LSTM[28] & ERP-LSTM(our) & Improvement \\ 
\midrule
\textbf{Cross-Subject} & 36.72\%      & 66.81\%      & \textbf{30.09\%}  \\ 
\midrule
\textbf{Within-Subject}  \\
Subject 1       & 35.07\%      & 89.06\%      & \textbf{53.99\%}  \\
Subject 2       & 35.30\%      & 60.94\%      & \textbf{25.64\% }          \\
Subject 3       & 45.25\%      & 71.88\%      & \textbf{26.63\% }          \\
Subject 4       & 35.88\%      & 50.00\%      & \textbf{14.12\% }          \\
Subject 5       & 48.03\%      & 65.62\%      & \textbf{17.59\%  }         \\
Subject 6       & 47.80\%      & 75.00\%      & \textbf{27.20\% }          \\
Subject 7       & 40.74\%      & 62.50\%      & \textbf{21.76\% }          \\
Subject 8       & 31.37\%      & 45.31\%      & \textbf{13.94\% }          \\
Subject 9       & 39.12\%      & 60.94\%      & \textbf{21.82\% }          \\
Subject 10      & 47.45\%      & 59.38\%      & \textbf{11.93\% }  \\

\bottomrule
\label{tabel 1}
\end{tabular}}
\end{table}
\end{center}

Our findings suggested that the representation feature space encoded from the extracted ERPs is more discriminative to classify image objects compared to that of the raw EEG. Also, we suppose that the critical information for object cognition of the brain signals did not miss during the averaging process. On the contrary, the extracted ERPs have retained the spatial and temporal feature that is related to the visual evoked potentials.

\subsection{Performance of Exemplar-Level Visual Classification}

Here, we further analysed the existing frameworks and our proposed ERP-LSTM framework at the exemplar image level. It removed the categories as the classification labels, and instead, it aims to identify a specific image as an exemplar. As shown in Table \ref{tabel 2}, we presented the existing two frameworks, Kaneshiro \cite{Kaneshiro2015} and EEG-LSTM \cite{spampinato2017}, to identify the exemplars with 72 classes across all 10 subjects. The findings showed that our proposed ERP-LSTM framework still could achieve the classification accuracy of 27.08\% at the exemplar level, which outperformed 14.46\% for Kaneshiro and 7.97\% for EEG-LSTM. We also attached the results of six-category level classification to get insights into the difference between easy (category) and hard (exemplar) modes.

\begin{table}[]
\caption{Performance of category- and exemplar-level visual classification}
\small
\setlength{\tabcolsep}{4mm}{
\begin{tabular}{cccc}
\toprule
Accuracy                 & Kaneshiro[18] & EEG-LSTM[28] & ERP-LSTM(our)  \\
\midrule
Categories (6 classes)  & 40.68\%   & 36.72\%  & \textbf{66.81\%}       \\
Exemplars (72 classes) & 14.46\%   & 7.97\%   & \textbf{27.08\%}        \\          
\bottomrule
\label{tabel 2}
\end{tabular}}
\end{table}
Thus, relative to the existing model, our work denoted that the representation feature decoded from the extracted ERPs is less confusion than raw EEG signals, which benefits to learn a more discriminative feature space for visual classification. Furthermore, our ERP-LSTM framework also achieved better performance than a recent work in 2020 \cite{renli2020} (in which the reported classification accuracy on 6 categories is 17.1\%), even if we used the different data source. This suggested that the LSTM network is capable to encode the ERPs to obtain a representative feature space, as the advantages of LSTM network on tackling temporal dynamics of time-series EEG signals. 

\section{Conclusion}

In this paper, we proposed an evoked potential-guided deep learning framework, called ERP-LSTM framework, for visual classification, which is separated into two stages: (1) extracting ERP sequences from multi-trial EEG segments; (2) a parallel LSTM network to encode a representation feature space for object categorisation as well as to classify EEG signal representations. Our proposed ERP-LSTM framework achieved better performance compared to existing frameworks both on the classification of 6 categories and 72 exemplar images. We believe our findings are presenting the feasibility to learn representational patterns of visual objects based on the recording of brain cortex activities, and an ERP-LSTM framework could learn characteristic features for visual classification.

%
%
%
\bibliographystyle{splncs04}
\bibliography{mybibliography.bib}

\begin{thebibliography}{10}
\providecommand{\url}[1]{\texttt{#1}}
\providecommand{\urlprefix}{URL }
\providecommand{\doi}[1]{https://doi.org/#1}

\bibitem{renli2020}
Ahmed, H., Wilbur, R.B., Bharadwaj, H.M., Siskind, J.M.: Object classification
  from randomized {EEG} trials. arXiv preprint arXiv:2004.06046  (2020)

\bibitem{bashivan2015learning}
Bashivan, P., Rish, I., Yeasin, M., Codella, N.: Learning representations from
  {EEG} with deep recurrent-convolutional neural networks. arXiv preprint
  arXiv:1511.06448  (2015)

\bibitem{Carlson_2011}
Carlson, T.A., Hogendoorn, H., Kanai, R., Mesik, J., Turret, J.: High temporal
  resolution decoding of object position and category. Journal of Vision
  \textbf{11}(10), ~9--9 (2011)

\bibitem{COX2003261}
Cox, D.D., Savoy, R.L.: Functional magnetic resonance imaging {(fMRI)}“brain
  reading”: detecting and classifying distributed patterns of fmri activity
  in human visual cortex. Neuroimage  \textbf{19}(2),  261--270 (2003)

\bibitem{Beeck10111}
De~Beeck, H.P.O., Torfs, K., Wagemans, J.: Perceived shape similarity among
  unfamiliar objects and the organization of the human object vision pathway.
  Journal of Neuroscience  \textbf{28}(40),  10111--10123 (2008)

\bibitem{gazzaniga2006cognitive}
Gazzaniga, M., Ivry, R., Mangun, G.: Cognitive neuroscience: The biology of the
  mind 3rd ed., ch. 6 (2008)

\bibitem{NIPS2018_7982}
Geirhos, R., Temme, C.R., Rauber, J., Sch{\"u}tt, H.H., Bethge, M., Wichmann,
  F.A.: Generalisation in humans and deep neural networks. In: Advances in
  Neural Information Processing Systems (NIPS). pp. 7538--7550 (2018)

\bibitem{Girshick_2014_CVPR}
Girshick, R., Donahue, J., Darrell, T., Malik, J.: Rich feature hierarchies for
  accurate object detection and semantic segmentation. In: The IEEE Conference
  on Computer Vision and Pattern Recognition (CVPR) (2014)

\bibitem{Grill1988}
Grill-Spector, K., Kushnir, T., Hendler, T., Edelman, S., Itzchak, Y., Malach,
  R.: A sequence of object-processing stages revealed by {fMRI} in the human
  occipital lobe. Human Brain Mapping  \textbf{6}(4),  316--328 (1998)

\bibitem{HANSON2004156}
Hanson, S.J., Matsuka, T., Haxby, J.V.: Combinatorial codes in ventral temporal
  lobe for object recognition: Haxby (2001) revisited: is there a “face”
  area? Neuroimage  \textbf{23}(1),  156--166 (2004)

\bibitem{Haynes2005}
Haynes, J.D., Rees, G.: Predicting the orientation of invisible stimuli from
  activity in human primary visual cortex. Nature Neuroscience  \textbf{8}(5),
  686--691 (2005)

\bibitem{Haynes2006Decoding}
Haynes, J.D., Rees, G.: Decoding mental states from brain activity in humans.
  Nature Reviews Neuroscience  \textbf{7}(7),  523--534 (2006)

\bibitem{He_2017_ICCV}
He, K., Gkioxari, G., Dollar, P., Girshick, R.: Mask {R-CNN}. In: The IEEE
  International Conference on Computer Vision (ICCV) (2017)

\bibitem{10.1007/978-3-319-10578-9_23}
He, K., Zhang, X., Ren, S., Sun, J.: Spatial pyramid pooling in deep
  convolutional networks for visual recognition. IEEE Transactions on Pattern
  Analysis and Machine Intelligence  \textbf{37}(9),  1904--1916 (2015)

\bibitem{He_2019_CVPR}
He, T., Zhang, Z., Zhang, H., Zhang, Z., Xie, J., Li, M.: Bag of tricks for
  image classification with convolutional neural networks. In: Proceedings of
  the IEEE Conference on Computer Vision and Pattern Recognition (CVPR). pp.
  558--567 (2019)

\bibitem{8260656}
Hosseini, H., Xiao, B., Jaiswal, M., Poovendran, R.: On the limitation of
  convolutional neural networks in recognizing negative images. In: 2017 16th
  IEEE International Conference on Machine Learning and Applications (ICMLA).
  pp. 352--358. IEEE (2017)

\bibitem{Ishai9379}
Ishai, A., Ungerleider, L.G., Martin, A., Schouten, J.L., Haxby, J.V.:
  Distributed representation of objects in the human ventral visual pathway.
  Proceedings of the National Academy of Sciences  \textbf{96}(16),  9379--9384
  (1999)

\bibitem{Kaneshiro2015}
Kaneshiro, B., Guimaraes, M.P., Kim, H.S., Norcia, A.M., Suppes, P.: A
  representational similarity analysis of the dynamics of object processing
  using single-trial {EEG} classification. Plos One  \textbf{10}(8),  e0135697
  (2015)

\bibitem{karpathy2016cs231n}
Karpathy, A., et~al.: Cs231n convolutional neural networks for visual
  recognition. Neural Networks  \textbf{1}, ~1 (2016)

\bibitem{renli2018}
Li, R., Johansen, J.S., Ahmed, H., Ilyevsky, T.V., Wilbur, R.B., Bharadwaj,
  H.M., Siskind, J.M.: Training on the test set? {An} analysis of {Spampinato}
  et al. [31]. arXiv preprint arXiv:1812.07697  (2018)

\bibitem{NORMAN2006424}
Norman, K.A., Polyn, S.M., Detre, G.J., Haxby, J.V.: Beyond mind-reading:
  multi-voxel pattern analysis of {fMRI} data. Trends in Cognitive Sciences
  \textbf{10}(9),  424 -- 430 (2006)

\bibitem{DEPASCALIS2004295}
Pascalis], V.D.: Chapter 16 - on the psychophysiology of extraversion. In:
  Stelmack, R.M. (ed.) On the Psychobiology of Personality, pp. 295 -- 327.
  Elsevier, Oxford (2004)

\bibitem{Peelen2007The}
Peelen, M.V., Downing, P.E.: The neural basis of visual body perception. Nature
  Reviews Neuroscience  \textbf{8}(8),  636--648 (2007)

\bibitem{philiastides2006neural}
Philiastides, M.G., Ratcliff, R., Sajda, P.: Neural representation of task
  difficulty and decision making during perceptual categorization: a timing
  diagram. Journal of Neuroscience  \textbf{26}(35),  8965--8975 (2006)

\bibitem{Philiastidesbhi130}
Philiastides, M.G., Sajda, P.: {Temporal Characterization of the Neural
  Correlates of Perceptual Decision Making in the Human Brain}. Cerebral Cortex
   \textbf{16}(4),  509--518 (2005)

\bibitem{qin2016classifying}
Qin, Y., Zhan, Y., Wang, C., Zhang, J., Yao, L., Guo, X., Wu, X., Hu, B.:
  Classifying four-category visual objects using multiple {ERP} components in
  single-trial {ERP}. Cognitive Neurodynamics  \textbf{10}(4),  275--285 (2016)

\bibitem{Simanova_2010}
Simanova, I., Van~Gerven, M., Oostenveld, R., Hagoort, P.: Identifying object
  categories from event-related {EEG}: toward decoding of conceptual
  representations. Plos One  \textbf{5}(12),  e14465 (2010)

\bibitem{spampinato2017}
Spampinato, C., Palazzo, S., Kavasidis, I., Giordano, D., Souly, N., Shah, M.:
  Deep learning human mind for automated visual classification. In: Proceedings
  of the IEEE Conference on Computer Vision and Pattern Recognition (CVPR). pp.
  6809--6817 (2017)

\bibitem{stober2015deep}
Stober, S., Sternin, A., Owen, A.M., Grahn, J.A.: Deep feature learning for
  {EEG} recordings. arXiv preprint arXiv:1511.04306  (2015)

\bibitem{Wang_2012}
Wang, C., Xiong, S., Hu, X., Yao, L., Zhang, J.: Combining features from {ERP}
  components in single-trial {EEG} for discriminating four-category visual
  objects. Journal of Neural Engineering  \textbf{9}(5),  056013 (2012)

\end{thebibliography}
%

\end{document}